\documentclass[11pt]{article}
\usepackage{amssymb}
\usepackage[preprint]{acl}

\usepackage{times}
\usepackage{latexsym}

\usepackage[T1]{fontenc}

\usepackage{pifont}        
\usepackage{color}

\usepackage[table]{xcolor}
\usepackage{arydshln} 

\usepackage{colortbl}  
\usepackage{booktabs}  
\usepackage{xcolor}    
\usepackage{microtype}

\definecolor{colConv}{RGB}{231, 245, 246}
\definecolor{colObv}{RGB}{224, 223, 239}
\definecolor{colContra}{RGB}{241, 232, 242}
\definecolor{colForm}{RGB}{242, 243, 244}
\definecolor{colAvg}{RGB}{239, 223, 212}
\usepackage{inconsolata}
\usepackage{makecell}
\usepackage{graphicx}
\usepackage{amsmath}
\usepackage{multirow}
\usepackage{amssymb}
\usepackage{booktabs}
\usepackage[normalem]{ulem}
\usepackage{graphicx}
\usepackage{subcaption}

\usepackage[utf8]{inputenc}

\usepackage{microtype}

\usepackage{inconsolata}

\usepackage{graphicx}

\title{SEA-Eval: A Benchmark for Evaluating Self-Evolving Agents Beyond Episodic Assessment}

\author{
  \textbf{Sihang Jiang, Lipeng Ma, Zhonghua Hong, Keyi Wang,  Zhiyu Lu, Tengfei Wang,} \\
  \textbf{Shisong Chen, Jinghao Zhang, Tianjun Pan, Weijia Li, Jiaqing Liang, Yanghua Xiao} \\
  Fudan University
}

\begin{document}
\maketitle

\begin{abstract}
Current LLM-based agents demonstrate strong performance in episodic task execution but remain constrained by static toolsets and episodic amnesia, failing to accumulate experience across task boundaries. This paper formalizes the Self-Evolving Agent (SEA) from the perspective of digital embodiment and continuous cross-task evolution, introduces the Evolutionary Flywheel as its minimal sufficient architecture, and presents SEA-Eval---the first benchmark designed specifically for evaluating SEAs. Grounded in Flywheel theory, SEA-Eval establishes $SR$ and $T$ as primary metrics and, through sequential task stream design, is designed to quantify evolutionary gain, evolutionary stability, and implicit alignment convergence. Empirical evaluation reveals that, under comparable success rates, token consumption differs by up to 31.2$\times$ between frameworks on individual tasks, with divergent evolutionary trajectories emerging under sequential analysis---demonstrating that success rate alone creates a capability illusion and that the sequential convergence of $T$ is the key criterion for distinguishing genuine evolution from pseudo-evolution.
\end{abstract}

\section{Introduction}
Large Language Model (LLM) agents have emerged as a pivotal paradigm for translating generative intelligence into tangible productivity. Since the inception of tool-augmented reasoning, agents have demonstrated substantial industrial utility across domains such as autonomous software engineering, scientific discovery, and complex workflow orchestration. Recent advancements in system-native architectures, exemplified by frameworks like \textit{GenericAgent}\footnote{\url{https://github.com/lsdefine/GenericAgent}}~\cite{genericagent2025} and \textit{Claude Code}\footnote{\url{https://docs.anthropic.com/en/docs/claude-code}}, have propelled agents beyond the confines of text-based reasoning engines. By achieving deep integration with operating systems and native applications, these agents are transitioning toward a state of \textbf{digital embodiment}: no longer mere semantic processors, but functional actors capable of sustaining high-load industrial tasks directly upon the underlying digital infrastructure. This evolutionary trajectory points toward a new agentic paradigm: the \textbf{Self-Evolving Agent (SEA)}.

The core characteristics of an SEA are defined by two properties. First, \textbf{complex interaction in open-ended environments}: the agent operates directly upon raw GUI pixels, dynamic DOM trees, and low-level system interfaces, unconstrained by predefined API boundaries. Second, \textbf{continuous, multi-turn evolution}: by maintaining persistent memory across task boundaries, the agent transforms historical interaction experience into cognitive priors that reconfigure the execution of subsequent tasks. Architecturally, the SEA is centered on the \textbf{Evolutionary Flywheel}---a closed-loop mechanism formed by the symbiotic coupling of an Execution Hub and a Cognition Hub, operating through three sequential phases of execution, distillation, and augmented execution. When the flywheel operates effectively, execution overhead converges monotonically as task frequency increases; when it fails, the agent falls into \textbf{pseudo-evolution}, relying on zero-shot reasoning with execution overhead scaling linearly with task frequency.

However, the novel characteristics of SEAs expose a fundamental mismatch with existing evaluation frameworks. Current benchmarks adopt the isolated task episode as the basic unit of evaluation, and are built around the episodic paradigm in which memory is reset at every task boundary---making the assessment of cross-task knowledge accumulation and continuous capability growth structurally impossible. This mismatch manifests across two dimensions: sequential task stream designs are absent, making it structurally impossible to capture cross-task evolutionary dynamics; and success-rate-dominated static metrics cannot capture the dynamic convergence of execution efficiency as experience accumulates, rendering pseudo-evolution entirely invisible to existing evaluation frameworks~\citep{zheng2025lifelongagentbenchevaluatingllmagents, cai2026buildingselfevolvingagentsexperiencedriven, gao2026surveyselfevolvingagentswhat}. This produces a capability illusion in practice: two frameworks achieving comparable success rates may differ fundamentally in evolutionary quality, yet existing benchmarks cannot distinguish them.

To bridge this evaluative gap, we introduce \textbf{SEA-Eval}---the first evaluation benchmark\footnote{https://leaperovo.github.io/SEA-Eval/} designed for self-evolving agents. SEA-Eval evaluates agents across two complementary dimensions: \textbf{intra-task execution reliability}, assessing immediate execution capability under environmental uncertainty independent of prior experience; and \textbf{long-term evolutionary performance}, assessing evolutionary gain, evolutionary stability, and implicit alignment convergence across sequential task deployments. In terms of metrics, we establish $SR$ and $T$ as the primary evaluative primitives, a selection grounded directly in the Evolutionary Flywheel framework: $SR$ serves as the minimum reliability guarantee for task completion, while the sequential convergence trajectory of $T$ serves as the core signal for distinguishing genuine evolution from pseudo-evolution. The dataset comprises 32 atomic tasks organized by three difficulty tiers and systematically composed into three types of sequential streams---correlated, orthogonal, and implicit intent---under two noise conditions that decouple retrieval robustness from evolutionary capability as an independent control variable.

Based on this framework, we conduct an empirical study evaluating leading agent frameworks including \textit{OpenClaw} and \textit{GenericAgent}. The results reveal a striking \textbf{Evolutionary Bottleneck}. While the frameworks attain comparable success rates (85\%--96.7\%), their token consumption differs by an order of magnitude---on individual tasks, OpenClaw consumes up to 31.2$\times$ as many tokens as GenericAgent for the same task. More importantly, the two diverge sharply when the same task is executed repeatedly: GenericAgent exhibits genuine evolutionary characteristics, with its token consumption decreasing monotonically across repetitions (by roughly 68\% in our experiments) as experience is distilled into low-cost execution strategies. OpenClaw, by contrast, displays the hallmarks of pseudo-evolution: its token consumption barely decreases across repeated executions, fails to form a stable convergence trajectory, and even rises when unrelated tasks are interleaved, exposing the fragility of its reliance on shortcut memorization. This contrast demonstrates that success rate alone creates an illusion of capability, while the sequential convergence of token consumption is necessary to faithfully reveal an agent's evolutionary quality.

The contributions of this work are three-fold:
\begin{itemize}
    \item We are the first to formalize the Self-Evolving Agent (SEA) from the perspective of digital embodiment and continuous cross-task evolution, and the first to introduce the Evolutionary Flywheel as its minimal sufficient architecture. This formalization, for the first time, provides a theoretically grounded criterion for distinguishing genuine evolution from pseudo-evolution, and characterizes two failure modes---distillation failure and retrieval failure.

    \item We introduce SEA-Eval, the first benchmark designed specifically for evaluating SEAs. We are the first to establish $SR$ and $T$ as a Flywheel-grounded metric system, which jointly covers two evaluation dimensions (intra-task execution reliability and long-term evolutionary performance). The accompanying dataset comprises 32 atomic tasks organized across three difficulty tiers and systematically composed into three types of sequential evaluation units, providing sufficient coverage and task diversity to support precise and comprehensive evaluation.

    \item We reveal a significant evolutionary bottleneck among mainstream frameworks: under comparable success rates in static evaluation, token consumption differs by up to 31.2$\times$ on individual tasks, with divergent evolutionary trajectories emerging under sequential analysis, which further demonstrates the effectiveness of SEA-Eval.
\end{itemize}

\section{Related Work}

\subsection{LLM Agents and Episodic Amnesia}

The development of autonomous agents has progressed from restricted, rule-based systems to increasingly open-ended and adaptive entities~\citep{wang2024survey, gao2026surveyselfevolvingagentswhat, xi2025rise}. The emergence of LLMs has catalyzed a new generation of agents capable of interleaving reasoning, planning, and tool use, forming iterative reasoning--action loops that enable interaction with complex environments~\citep{yehudai2025survey, rahman2025llm}. However, current LLM-based agents remain fundamentally constrained by episodic amnesia. We extend this concept from cognitive science to the agent lifecycle: just as patients with episodic amnesia retain procedural skills yet cannot encode specific experiences into persistent memory, LLM agents demonstrate strong within-task reasoning yet reset their cognitive state at every task boundary, failing to transform interaction experience into reusable cross-task cognitive assets~\citep{luo2025large, du2026memory}. This limitation has been studied in the continual learning literature, where methods such as EWC~\citep{kirkpatrick2017overcoming} mitigate forgetting through parameter-level update mechanisms~\citep{cao2025dare}. However, these approaches define evolution at the algorithm level and presuppose specific update mechanisms, whereas SEA defines evolution at the architecture level without constraining implementation. SEA further operates over open-ended digital environments rather than closed task distributions, and grounds evolution in external memory rather than model parameters.

\subsection{Technical Foundations of Self-Evolving Agents}

Prior work has advanced the technical capabilities relevant to SEA along two lines corresponding to its dual-hub architecture. On the Execution Hub side, CREATOR~\citep{qian2023creator}, Voyager~\citep{wang2023voyager}, and ToolLLM~\citep{qin2023toolllm} have explored tool creation, skill library construction, and large-scale tool selection respectively, while embodied AI research~\citep{brohan2023rt, duan2022embodied} has extended the perception--action paradigm to physical and digital environments. On the Cognition Hub side, Reflexion~\citep{shinn2023reflexion}, ExpeL~\citep{zhao2024expelllmagentsexperiential}, and MemGPT~\citep{packer2023memgpt} have advanced memory system construction; CoT~\citep{wei2022cot} and ReAct~\citep{yao2022react} have improved planning and reasoning; and Self-Refine~\citep{madaan2023selfrefine} and CRITIC~\citep{gou2023critic} have strengthened self-reflection mechanisms.

Nevertheless, these works each advance isolated capabilities without recognizing that the underlying dimensions are structurally interdependent: memory quality depends on reflection, reflection signals derive from execution, and execution efficiency reflects memory effectiveness. No prior work has formalized this interdependence as a closed-loop system or identified the minimal sufficient conditions for instantiating the SEA definition. To the best of our knowledge, this paper provides the first formal characterization of this closed-loop structure.

\subsection{Limitations of Existing Evaluation Frameworks}

Because prior work treats these dimensions as independent techniques rather than a closed-loop system, existing benchmarks have not been designed to capture the quality of closed-loop evolution. Mainstream benchmarks including WebArena~\citep{zhou2024webarenarealisticwebenvironment}, OSWorld~\citep{xie2024osworldbenchmarkingmultimodalagents}, SWE-bench~\citep{jimenez2024swebenchlanguagemodelsresolve}, $\tau$-bench~\citep{yao2024taubenchbenchmarktoolagentuserinteraction}, and AgentBench~\citep{liu2025agentbenchevaluatingllmsagents} all adopt the single task episode as the basic evaluation unit, effectively measuring immediate execution competence while remaining blind to cross-task evolutionary dynamics. Recent works including LifelongAgentBench~\citep{zheng2025lifelongagentbenchevaluatingllmagents}, MetaTool~\citep{metatool}, HammerBench~\citep{hammerbench}, and EverMemBench~\citep{everarena} have begun to explore experience reuse and efficiency dimensions, but treat efficiency as a static single-task property rather than a dynamic trend across sequential task streams. If an agent's token consumption fails to converge after repeated execution of structurally similar tasks, this is the true signal of absent evolutionary capability, yet it remains entirely invisible under existing frameworks. Among available efficiency proxies, token consumption most faithfully reflects this dynamic: execution step count may remain stable even as per-step reasoning depth increases, and wall-clock latency introduces hardware-dependent variance that compromises cross-environment reproducibility.

This gap manifests across two dimensions that directly motivate SEA-Eval's design. First, on sequential design, SEA-Eval introduces persistent memory configurations and ordered task streams, replacing isolated episodes with cross-task behavioral trajectories and enabling for the first time the direct observation of cross-task knowledge accumulation and continuous capability growth. Second, on efficiency metrics, SEA-Eval tracks the convergence trajectory of token consumption across sequential task streams rather than static single-task measurements: if an agent's token consumption fails to converge after repeated execution of structurally similar tasks, this is the true signal of absent evolutionary capability---a signal that remains entirely invisible under existing frameworks~\citep{gao2026surveyselfevolvingagentswhat, cai2026buildingselfevolvingagentsexperiencedriven, zheng2025lifelongagentbenchevaluatingllmagents}.

\section{The SEA Paradigm}
\label{sec:sea}

\subsection{Foundational Characteristics}
\label{sec:characteristics}
We formally define a self-evolving agent as a tuple $\mathcal{SEA} = (E, \mathcal{A}, \mathcal{M}, \Phi, \Psi)$, where $E$ denotes an open-ended digital environment characterized by dynamic state spaces, heterogeneous interaction modalities, and the absence of pre-enumerated action constraints; $\mathcal{A}$ an open and extensible action space subject to autonomous expansion; $\mathcal{M} = \mathcal{M}_p \cup \mathcal{M}_e \cup \mathcal{M}_d$ a persistent stratified memory maintained across task boundaries, encompassing procedural knowledge $\mathcal{M}_p$ (reusable execution strategies), episodic knowledge $\mathcal{M}_e$ (historical trajectories and failure attributions), and declarative knowledge $\mathcal{M}_d$ (environmental constraints and user preference models); $\Phi: \tau \rightarrow \Delta\mathcal{M}$ an experience distillation function mapping raw interaction trajectories to structured memory updates; and $\Psi: \mathcal{M}_t \times \Delta\mathcal{M} \rightarrow \mathcal{M}_{t+1}$ a selective cross-task memory transfer function that determines which updates produced by $\Phi$ are committed to persistent memory. A detailed discussion of the functional boundary between $\Phi$ and $\Psi$ is provided in Appendix~\ref{appendix:phi_psi}.

This definition is grounded in two non-negotiable foundational characteristics.

\paragraph{Complex Interaction in Open-ended Environments.}
An SEA operates within open-ended digital ecosystems that are not constrained to application-specific domains or pre-enumerated action schemas. Unlike benchmarks that evaluate agents within simplified synthetic environments or sandboxes restricted to specific applications, the environments in which SEAs are deployed exhibit dynamic state spaces, heterogeneous interaction modalities, and persistent stochasticity that cannot be fully anticipated at design time. Under these conditions, the agent must engage directly with raw GUI pixels, dynamic DOM trees, and low-level system interfaces---a mode of operation we term \textbf{digital embodiment}: the capacity for unmediated, system-native interaction with the underlying computational substrate, unconstrained by predefined interface abstractions.

\paragraph{Continuous, Multi-turn Evolution.}
The second characteristic defines the temporal boundary of the SEA paradigm. While episodic agents reset $\mathcal{M}$ upon task completion, an SEA maintains state persistence across unbounded deployment cycles via $\Psi$. The outcomes of prior tasks---potentially occurring hours or days earlier---are retained in $\mathcal{M}$ and actively reconfigure the execution path of subsequent, structurally unrelated interactions. This property overcomes the episodic amnesia inherent in context-window-bounded systems and establishes cross-task behavioral continuity as a definitional requirement: an agent that resets its cognitive state at every task boundary is, by definition, not evolving, regardless of the sophistication of its within-task reasoning. In episodic agents, $\Phi$ and $\Psi$ degenerate to null operations---$\mathcal{M}$ is reset at every task boundary and $\mathcal{A}$ remains static. The SEA definition formally demarcates this boundary by requiring $\Phi$ and $\Psi$ to be non-trivial: any system in which these functions are effectively identity mappings is, by definition, an episodic agent regardless of its within-task reasoning sophistication.

\begin{figure}[t]
    \centering
    \includegraphics[width=\columnwidth]{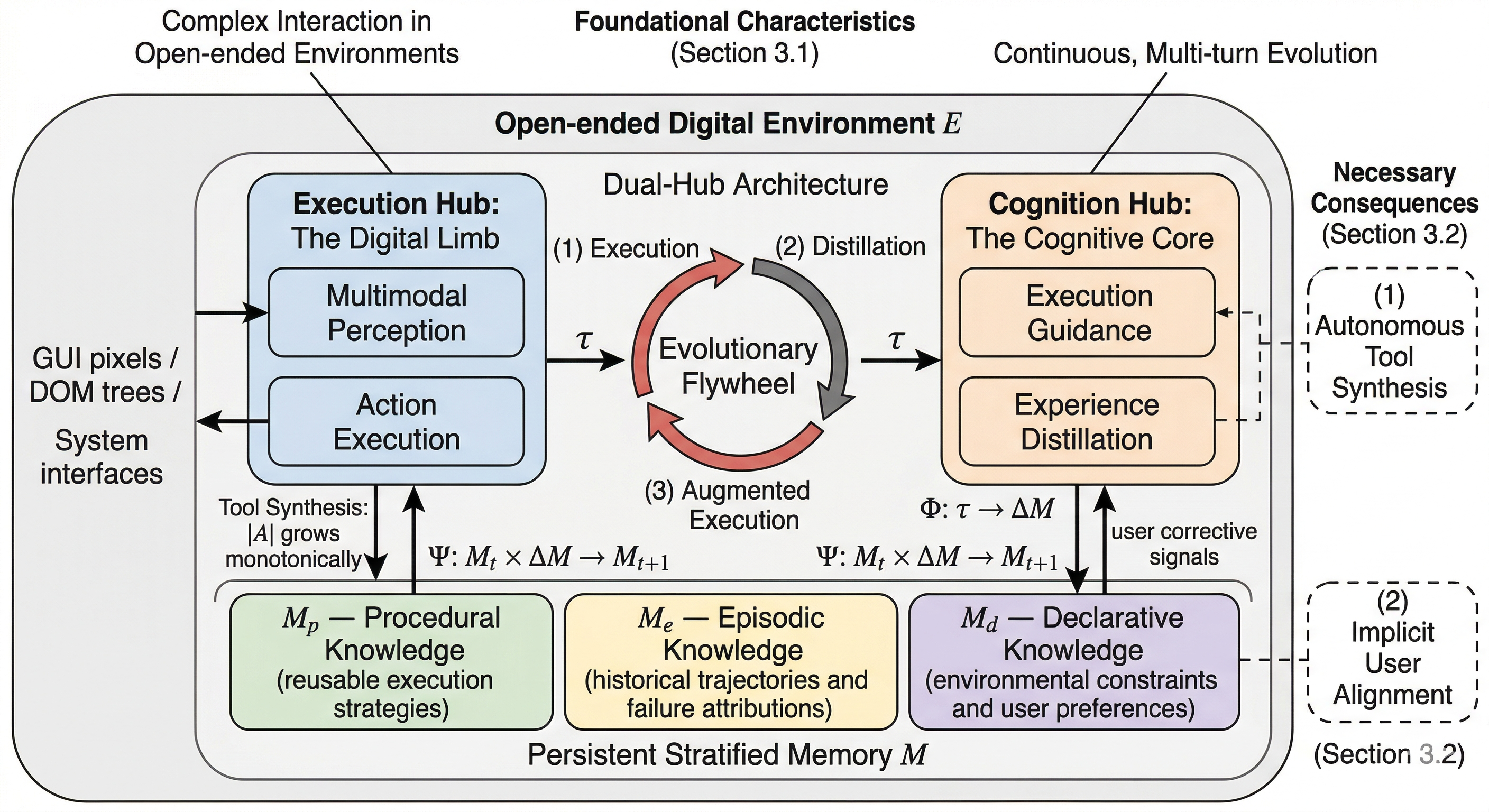}
    \caption{The Self-Evolving Agent (SEA) framework.}
    \label{sea_framework}
\end{figure}

\subsection{Necessary Consequences of the SEA Characteristics}
\label{sec:consequences}
The two foundational characteristics defined above are not merely descriptive---they entail necessary behavioral consequences that any SEA must exhibit in open-ended deployment. We identify two such derived properties, each following by logical necessity from one of the foundational characteristics.

\paragraph{Autonomous Tool Synthesis.}
The first characteristic---complex interaction in open-ended environments---directly entails the necessity of autonomous tool synthesis. In environments where action constraints are not pre-enumerated, pre-defined tool schemas will inevitably encounter unhandled states: the agent will face interaction requirements that no existing tool in $\mathcal{A}$ is equipped to satisfy. Under these conditions, the agent must autonomously synthesize new functional primitives, generating and registering validated tools through whatever mechanisms are available to it. This process directly implements the expansion of $\mathcal{A}$: the cardinality $|\mathcal{A}|$ grows monotonically over deployment time as a function of environmental engagement rather than pre-programmed capability. Autonomous tool synthesis is therefore not an optional capability enhancement but a structural necessity imposed by the open-ended nature of $E$.

\paragraph{Implicit User Alignment.}
The second characteristic---continuous, multi-turn evolution---directly entails the progressive alignment of the agent's behavior with user-specific preferences. Over extended deployment cycles, the SEA accumulates interaction data with a specific user, including corrective signals such as rollbacks, revisions, and explicit feedback. These signals constitute a high-dimensional reward source that, through repeated application of $\Phi$, drives the progressive encoding of user-specific workflows and implicit preferences into $\mathcal{M}_d$. The result is an alignment process in which the agent's execution strategies increasingly mirror the user's latent intent---not through one-time calibration, but through continuous distillation of accumulated interaction friction. This emergent alignment is a necessary consequence of sustained multi-turn evolution with a consistent user, rather than a separately engineered component.

\subsection{The Evolutionary Flywheel}

To instantiate the characteristics defined in Section~\ref{sec:characteristics}, an SEA requires a structural backbone capable of closing the loop between raw environmental interaction and the persistent updates to $\mathcal{M}$ governed by $\Phi$ and $\Psi$. We argue that this structural backbone is not an arbitrary design choice but a \textbf{minimal sufficient architecture}: any system satisfying the $\mathcal{SEA}$ definition must implement mechanisms functionally equivalent to the following two hubs and their coupling, in the sense that removing either hub or their interaction would violate at least one of the foundational characteristics defined in Section~\ref{sec:characteristics}. The dual-hub architecture described below therefore represents a structural reference derived from the $\mathcal{SEA}$ definition, rather than a novel algorithmic contribution in itself, as shown in Figure~\ref{sea_framework}.

\subsubsection{Execution Hub: The Digital Limb}

The Execution Hub serves as the agent's sensory-motor interface with $E$, implementing the operational requirements of digital embodiment through two functional components.

\paragraph{Multimodal Perception.}
The hub continuously acquires heterogeneous signals from the environment---including raw screen visuals, DOM structures, system logs, and terminal outputs---and transforms them into environmental representations available for decision-making. This perceptual capacity is the prerequisite for system-native interaction: without the ability to acquire environmental feedback across modalities, the agent cannot operate beyond the confines of predefined API interfaces.

\paragraph{Action Execution.}
The hub executes actions within $\mathcal{A}$, and autonomously synthesizes new tools when existing ones prove insufficient for the current environmental demands, registering them as persistent tools for future use.

\subsubsection{Cognition Hub: The Cognitive Core}

The Cognition Hub implements the $\Phi$ and $\Psi$ functions that constitute the SEA's evolutionary capacity. It operates across two temporal modes: guiding the Execution Hub during task execution, and distilling accumulated experience into persistent memory updates after task completion.

\paragraph{Execution Guidance.}
During task execution, the Cognition Hub provides decision support to the Execution Hub by maintaining a representation of the current task state, retrieved context from $\mathcal{M}$, and a continuously updated execution plan. This guidance integrates prior experience from $\mathcal{M}_p$ and $\mathcal{M}_d$ into the current execution context, enabling the agent to draw on consolidated strategies rather than reasoning from scratch at each decision step. When execution anomalies are detected, the Cognition Hub revises the current plan, maintaining closed-loop execution stability under environmental perturbation.

\paragraph{Experience Distillation.}
Following task completion, the Cognition Hub instantiates $\Phi$, distilling from the raw interaction trajectory $\tau$ the experience most beneficial to future execution---including reusable execution strategies, failure attributions, and user preference signals---and organizing these into a structured update $\Delta\mathcal{M}$. These updates are then passed to $\Psi$, which selectively integrates knowledge assessed as sufficiently generalizable into the persistent memory state $\mathcal{M}_{t+1}$, suppressing instance-specific noise to preserve evolutionary stability.

\subsubsection{The Evolutionary Flywheel and Its Failure Modes}
\label{sec:flywheel}
The symbiotic coupling of the Execution Hub and Cognition Hub constitutes the \textbf{Evolutionary Flywheel}: a closed-loop mechanism through which raw environmental interaction is progressively transformed into persistent cognitive assets. The flywheel operates through three sequential phases:

\begin{enumerate}
    \item \textbf{Execution}: The Execution Hub engages $E$, generating a raw interaction trajectory $\tau$ that captures the full sequence of perceptual inputs, actions, environmental responses, and error recovery events.
    \item \textbf{Distillation}: The Cognition Hub applies $\Phi$ to $\tau$, distilling valuable experience into structured updates $\Delta\mathcal{M}$, which are then selectively committed to $\mathcal{M}_{t+1}$ via $\Psi$.
    \item \textbf{Augmented Execution}: In subsequent tasks, the Execution Hub retrieves relevant entries from $\mathcal{M}$---particularly consolidated strategies from $\mathcal{M}_p$---replacing exhaustive zero-shot reasoning with experience-guided execution. The resulting trajectory is of higher quality and lower cost than its predecessor, producing richer material for the next distillation cycle.
\end{enumerate}

These three phases form a self-reinforcing loop: each execution cycle produces experience that, once distilled, reduces the cost and increases the quality of future execution. The flywheel generates a net evolutionary gain when the reduction in execution cost achieved through experience-guided execution exceeds the overhead introduced by distillation and retrieval. This condition is met when $\Phi$ successfully abstracts generalizable strategies from instance-specific trajectories---transforming task-level successes into reusable cross-task knowledge.

The flywheel can degrade into \textbf{pseudo-evolution} under two theoretically distinct failure modes. In \textbf{distillation failure}, $\Phi$ fails to distill generalizable experience from the trajectory, producing memory updates that cannot transfer beyond the specific instance and leaving $\mathcal{M}_p$ inert despite apparent task success. In \textbf{retrieval failure}, the Execution Hub cannot identify structural similarity between a new task and existing entries in $\mathcal{M}$---particularly under environmental drift---and defaults to exhaustive zero-shot reasoning, bypassing the flywheel entirely. Critically, both failure modes produce an identical behavioral signature: execution cost fails to decrease across sequential task variants and scales linearly with task frequency rather than converging. A detailed diagnostic discussion of how these two failure modes can be distinguished through auxiliary execution traces is provided in Appendix~\ref{appendix:failure}.

\begin{figure*}[t]
    \centering
    \includegraphics[width=0.9\textwidth]{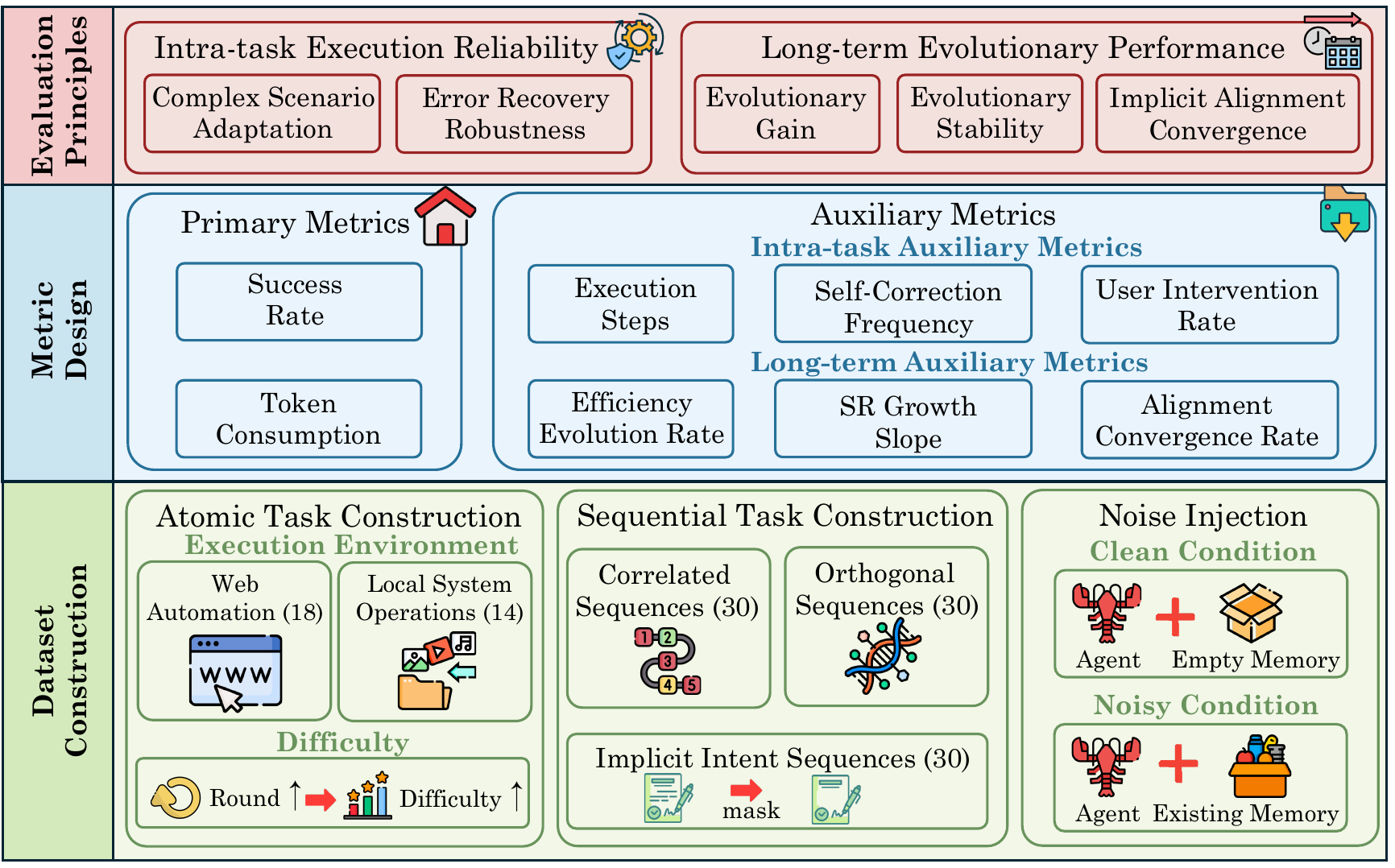}
    \caption{The SEA-Eval framework design pipeline across three layers.}
    \label{fig:sea_eval_design}
\end{figure*}

\section{SEA-Eval: Evaluation Framework and Benchmark Design}

\subsection{Evaluation Principles}
\label{sec:principles}

The theoretical analysis in Section~\ref{sec:sea} establishes that \textbf{genuine evolution} in SEAs produces a distinguishable behavioral signature: as the Evolutionary Flywheel operates above the distillation-yield threshold, execution cost \textbf{converges monotonically} across sequential task variants while task success is maintained. Conversely, \textbf{pseudo-evolution}---arising from distillation or retrieval failure---produces no such convergence, with execution cost scaling \textbf{linearly} with task frequency regardless of accumulated experience. These predictions jointly motivate an evaluation design that must simultaneously track \textbf{task success} and \textbf{execution overhead} across temporally connected task sequences, rather than assessing isolated episodic tasks. SEA-Eval instantiates this requirement across two complementary dimensions, as shown in Figure~\ref{fig:sea_eval_design}.

\subsubsection{Intra-task Execution Reliability}

This dimension directly instantiates the first foundational characteristic of the SEA paradigm: \textbf{complex interaction} in open-ended environments. It evaluates the agent's capacity to navigate high-complexity, non-stationary tasks without relying on prior accumulated experience, assessing \textbf{immediate execution capability} under environmental uncertainty independent of long-term memory evolution. Intra-task execution reliability encompasses two aspects: \textbf{complex scenario adaptation} and \textbf{error recovery robustness}. A well-functioning SEA should sustain coherent task progress across multi-step decision chains and long-horizon reasoning sequences despite the dynamic and unpredictable nature of the deployment environment; when encountering execution failures or unexpected environmental perturbations, it should maintain closed-loop execution through self-reflection and self-correction, avoiding logical deadlocks or cascading system failures.

\subsubsection{Long-term Evolutionary Performance}

This dimension directly instantiates the second foundational characteristic of the SEA paradigm: \textbf{continuous, multi-turn evolution}. It assesses whether the agent's performance systematically improves over successive task deployments and whether these improvements remain stable across task boundaries. Long-term evolutionary performance encompasses three aspects: \textbf{evolutionary gain}, \textbf{evolutionary stability}, and \textbf{implicit alignment convergence}. A well-functioning SEA should exhibit all three. In terms of evolutionary gain, a well-functioning SEA should demonstrate measurable performance improvement as task frequency increases, with execution overhead \textbf{continuously decreasing}---this convergence trend constitutes the empirical signature of a functioning Evolutionary Flywheel. In terms of evolutionary stability, a well-functioning SEA should ensure that capabilities acquired through new tasks do \textbf{not} adversely affect ongoing execution performance, and that the continuous expansion of $\mathcal{M}$ does not interfere with the effective retrieval of existing procedural knowledge. In terms of implicit alignment convergence, a well-functioning SEA should exhibit a \textbf{monotonically decreasing} user intervention rate over extended deployment with a consistent user, reflecting the progressive internalization of user-specific preferences into $\mathcal{M}_d$---directly instantiating the implicit user alignment property identified in Section~\ref{sec:consequences} as a necessary consequence of continuous multi-turn evolution.

\subsection{Metric Design}
\label{sec:metrics}
 
SEA-Eval adopts Success Rate ($SR$) and Token Consumption ($T$) as its two primary evaluative primitives, a selection motivated directly by the Evolutionary Flywheel framework and the requirement to identify pseudo-evolution. $SR$ serves as the minimum reliability guarantee for task completion, establishing whether an agent possesses the baseline execution capability required for deployment. $T$ serves as the core signal for detecting whether the Evolutionary Flywheel is genuinely operating: when an agent successfully distills historical experience into reusable strategies, execution overhead should decrease continuously as task frequency increases; if $T$ fails to converge across sequential tasks, this indicates that the agent has fallen into pseudo-evolution regardless of its $SR$ performance. Beyond $SR$ and $T$, SEA-Eval defines a set of auxiliary metrics that provide finer-grained diagnostic analysis of the evolutionary process.
 
\subsubsection{Primary Metrics}
 
\paragraph{Success Rate ($SR$).}
$SR$ quantifies the accuracy of goal achievement, serving as the baseline measure of task execution capability and the minimum reliability guarantee for deployment viability.
 
\paragraph{Token Consumption ($T$).}
$T$ measures the computational overhead of task execution and serves as the core signal for detecting the operational state of the Evolutionary Flywheel. When an agent achieves genuine evolution, experience-guided execution progressively replaces exhaustive zero-shot reasoning, and $T$ should exhibit a monotonically convergent trend; when the flywheel fails, $T$ scales linearly with task frequency, exposing the agent's pseudo-evolutionary nature.
 
\subsubsection{Intra-task Auxiliary Metrics}
 
Within a single task execution cycle, the following metrics provide diagnostic granularity beyond the primary $SR$ and $T$ measurements.
 
\paragraph{Execution Steps ($S$).}
$S$ records the number of decision-making steps required to complete a task, providing a coarse-grained indicator of action planning efficiency that complements $T$ in characterizing intra-task execution paths.
 
\paragraph{Self-Correction Frequency ($N_{\text{fix}}$).}
$N_{\text{fix}}$ records the frequency of autonomous error identification and recovery within a single task, reflecting the agent's real-time resilience under environmental perturbation and complementing $SR$ as a measure of execution robustness.
 
\paragraph{User Intervention Rate ($N_{\text{undo}}$).}
$N_{\text{undo}}$ records the frequency of manual corrections, command rollbacks, or user-initiated interventions during task execution, quantifying the degree of inconsistency between the agent's execution logic and user intent. Its longitudinal behavior across task sequences forms the basis for the alignment convergence metric defined below.
 
\subsubsection{Long-term Auxiliary Metrics}
 
Long-term evolution is assessed by constructing a time series over sequential task deployments and applying dynamic analysis to characterize evolutionary trajectories.
 
\paragraph{Efficiency Evolution Rate ($K_{\text{eff}}$).}
$K_{\text{eff}}$ characterizes the trend of token consumption change across sequential task variants:
\begin{equation}
K_{\text{eff}}^{(k)} = \frac{T_{k+1} - T_k}{\Delta N}
\end{equation}
A $K_{\text{eff}}^{(k)}$ sequence that converges monotonically toward zero from the negative side indicates successful experience consolidation, with the agent progressively replacing exhaustive reasoning with experience-guided execution; a flat or erratic sequence indicates evolutionary stagnation.
 
\paragraph{SR Growth Slope ($K_{SR}$).}
$K_{SR}$ measures the trend of $SR$ improvement along the task sequence:
\begin{equation}
K_{SR}^{(k)} = \frac{SR_{k+1} - SR_k}{\Delta N}
\end{equation}
$K_{SR}$ characterizes the agent's learning capacity and growth ceiling. Jointly analyzed with $K_{\text{eff}}$, the two metrics form the combined criterion for identifying pseudo-evolution: when both simultaneously approach zero, the agent completes tasks without reducing execution overhead or improving reliability across sequential variants, confirming evolutionary stagnation as the empirical manifestation of flywheel failure.
 
\paragraph{Alignment Convergence Rate ($K_{\text{align}}$).}
$K_{\text{align}}$ measures the decreasing trend of user intervention frequency as the task sequence progresses:
\begin{equation}
K_{\text{align}}^{(k)} = \frac{\bar{N}_{\text{undo},k+1} - \bar{N}_{\text{undo},k}}{\Delta N}
\end{equation}
where $\bar{N}_{\text{undo},k}$ denotes the sliding-window average of the user intervention rate at step $k$. A $K_{\text{align}}^{(k)}$ sequence converging toward zero from the negative side indicates that the agent is successfully internalizing user-specific preferences into $\mathcal{M}_d$, directly instantiating the implicit alignment convergence dimension defined in Section~\ref{sec:principles}.
 
\paragraph{From Framework-level Definitions to Experimental Aggregation.}
The step-wise rates defined above---$K_{\text{eff}}^{(k)}$, $K_{SR}^{(k)}$, and $K_{\text{align}}^{(k)}$---constitute the framework-level primitives that characterize the evolutionary trajectory at the granularity of a single sequence transition, i.e., the increment between two adjacent steps. In the experiments, these primitives can be adapted according to the specific evaluation target---for instance, to observe the change induced after interference is inserted, or to construct a learning-retention measure that reflects the balance between acquiring new knowledge and preserving previously consolidated knowledge under interference.

\subsection{Dataset Design}
 
To evaluate both the static execution capability and the dynamic evolutionary capability of SEAs, the dataset is constructed from atomic tasks as the fundamental building blocks, which are then systematically composed into task sequences as the primary evaluation units. A noise injection mechanism is further introduced as an independent control variable to assess retrieval robustness under realistic memory congestion conditions.
 
\subsubsection{Atomic Task Construction}
 
To evaluate the foundational execution capability of agents on individual tasks, we first construct a set of atomic tasks as the basic units of the benchmark.
 
\paragraph{Scenario and Capability Coverage}
To ensure comprehensive coverage of digital embodiment scenarios, we select two categories of tasks: \textbf{Web Automation} and \textbf{Local System Operations}. The former encompasses web browsing and manipulation at the browser layer; the latter encompasses interactions at the operating system's native layer, including file management, script execution, system configuration, and native desktop application operations. The benchmark imposes no predefined tool interfaces; agents are free to complete tasks via GUI manipulation, command-line alternatives, or API calls, and the autonomous selection of interaction pathways is itself part of what is being evaluated. The union of two scenario categories achieves logical completeness over the execution path space of information-processing digital embodiment tasks, as any digital task can necessarily be completed via either a local native path or a web access path.
 
\paragraph{Three-tier Difficulty System}
To ensure that the benchmark is discriminative across agents of varying capability levels, atomic tasks are organized into three difficulty tiers. A single-difficulty task set cannot simultaneously satisfy the dual requirements of baseline stability and evolutionary discriminability: tasks that are too simple approach a success-rate ceiling that precludes the observation of evolutionary gain, while tasks that are too difficult produce unstable baseline success rates that render evolutionary signals uninterpretable.
 
\begin{itemize}
    \item \textbf{Easy}: Standard atomic operations with clear execution paths, requiring basic tool invocation and state tracking. These tasks establish the baseline capability reference and provide the statistical stability necessary for long-horizon evolutionary analysis.
    \item \textbf{Normal}: Tasks that introduce multiple rounds of complex environmental judgment into the execution process, involving conditional branching and intermediate state uncertainty. These tasks constitute the primary evaluation objects of the benchmark and provide the main source of discriminability for evolutionary capability assessment.
    \item \textbf{Hard}: Tasks involving long-range dependencies and extensive complex retrieval, requiring the agent to integrate information across steps and perform deep reasoning. These tasks are positioned as stress tests to probe the upper boundary of intra-task execution reliability under extreme complexity.
\end{itemize}
 
\paragraph{Variable Slot Design}
To support the systematic generation of structurally equivalent task instances within sequences, each atomic task contains one to two replaceable variable slots---such as retrieval targets, destination websites, or file names. Replacing a slot changes the specific execution object without altering the task's underlying execution logic or difficulty tier; all variants are reviewed by domain experts and accompanied by concrete examples to ensure equivalence. This design also prevents agents from reducing execution overhead by memorizing specific answers rather than learning generalizable strategies, ensuring that efficiency changes observed across sequences faithfully reflect evolutionary capability rather than memorization effects.
 \subsubsection{Sequential Task Construction}

To evaluate the dynamic evolutionary capability of agents, atomic tasks are systematically composed into task sequences, which serve as the primary evaluation units of the benchmark. A well-functioning SEA is expected to exhibit three properties: it should learn positively from repeated experience and progressively improve execution efficiency; it should resist negative interference from unrelated tasks; and its learning should manifest as transferable, generalizable strategies rather than rote memorization of specific execution paths. Our sequence design is organized around these three properties.

Let $A$ denote a target atomic task, and let $A_1, A_2, A_3$ denote distinct variable-slot instantiations of $A$ that share identical execution logic but differ only in concrete operands. We take $A_1$ as the reference instance and write $R_1, R_2, R_3$ for three consecutive repeated executions of $A_1$; $B, C, D$ denote unrelated tasks drawn from different scenarios and capability dimensions. On this basis, we construct two types of sequences:

\begin{itemize}
    \item \textbf{Correlated Sequences} ($R_1 \!\to\! R_2 \!\to\! R_3 \!\to\! A_2 \!\to\! A_3$): The leading segment repeats the identical instance $A_1$ to examine whether the agent learns from identical experience and progressively reduces execution overhead; the trailing segment switches to the similar instances $A_2, A_3$ to examine whether the acquired strategies generalize to new instantiations of the same task. If efficiency gains appear only on the identical repetitions but fail to carry over to the similar instances, the agent is relying on rote memorization rather than strategy generalization.

    \item \textbf{Orthogonal Sequences}: Unrelated interference tasks ($B, C, D$) are inserted between executions of the target task to examine whether previously acquired skills degrade under cross-domain interference. Two sub-conditions are constructed: \textbf{Orth-Same} ($A_1 \!\to\! B \!\to\! C \!\to\! D \!\to\! A_1$) re-executes the identical instance after interference, measuring pure retention robustness; \textbf{Orth-Similar} ($A_1 \!\to\! B \!\to\! C \!\to\! D \!\to\! A_2$) executes a different instance after interference, further examining whether generalization ability survives distraction.
\end{itemize}

The sequence configuration is summarized in Table~\ref{tab:dataset}. All sequences are set to length~5, balancing trend observability against evaluation overhead. After public release, the benchmark can be extended by replacing variable slots to generate new instances, designing additional atomic tasks, or adjusting sequence lengths according to available computational budgets.

\begin{table}[h]
\centering
\caption{Sequence configuration of SEA-Eval.}
\label{tab:dataset}
\scriptsize
\begin{tabular}{llc}
\toprule
\textbf{Type} & \textbf{Structure} & \textbf{\# Easy \& Normal} \\
\midrule
Correlated & $R_1 \!\to\! R_2 \!\to\! R_3 \!\to\! A_2 \!\to\! A_3$  & 12 / 18 \\
Orth-Same & $A_1 \!\to\! B \!\to\! C \!\to\! D \!\to\! A_1$  & 12 / 18 \\
Orth-Similar & $A_1 \!\to\! B \!\to\! C \!\to\! D \!\to\! A_2$ & 12 / 18 \\
\bottomrule
\end{tabular}
\end{table}
 
\subsubsection{Noise Injection Mechanism}
 
To evaluate retrieval robustness under realistic deployment conditions, a noise injection mechanism is introduced as an independent control variable. In real-world deployment, an agent's memory store $\mathcal{M}$ expands continuously over time, and degradation in retrieval precision under memory congestion is an unavoidable operational challenge that directly corresponds to the retrieval failure mode analyzed in Section~\ref{sec:flywheel}. By orthogonally combining noise conditions with sequence types, this mechanism decouples retrieval robustness from evolutionary capability and quantifies it as an independent dimension.
 
\begin{itemize}
    \item \textbf{Clean Condition}: The agent begins each sequence from an empty memory state, carrying no pre-acquired skills. This condition serves as the baseline, eliminating the influence of prior knowledge and ensuring that sequence evaluation results faithfully reflect the agent's ability to learn from scratch.
    \item \textbf{Noisy Condition}: The agent is preloaded with 20 randomly selected skills that have no scenario overlap with the current task sequence, simulating a memory congestion state representative of real-world deployment. The number of preloaded skills substantially exceeds the number required for any single sequence, thereby constituting an effective retrieval interference. This condition is used to assess whether skill bloat induces retrieval failure and whether pre-acquired knowledge interferes with the evolutionary trajectory of the current sequence.
\end{itemize}
 
\subsubsection{Dataset Statistics}

The complete SEA-Eval dataset comprises 32 atomic tasks---12 Easy, 18 Normal, and 2 Hard---distributed across Web Automation and Local System Operations. A total of 92 sequences are constructed from these atomic tasks: 32 correlated sequences, 30 orthogonal sequences and 30 implicit intent sequences. All tasks are designed by human experts based on publicly accessible resources, ensuring baseline solvability and ground-truth verifiability. No user privacy data is involved in any part of the dataset.
\begin{figure*}[t]
\centering
\includegraphics[width=0.9\linewidth]{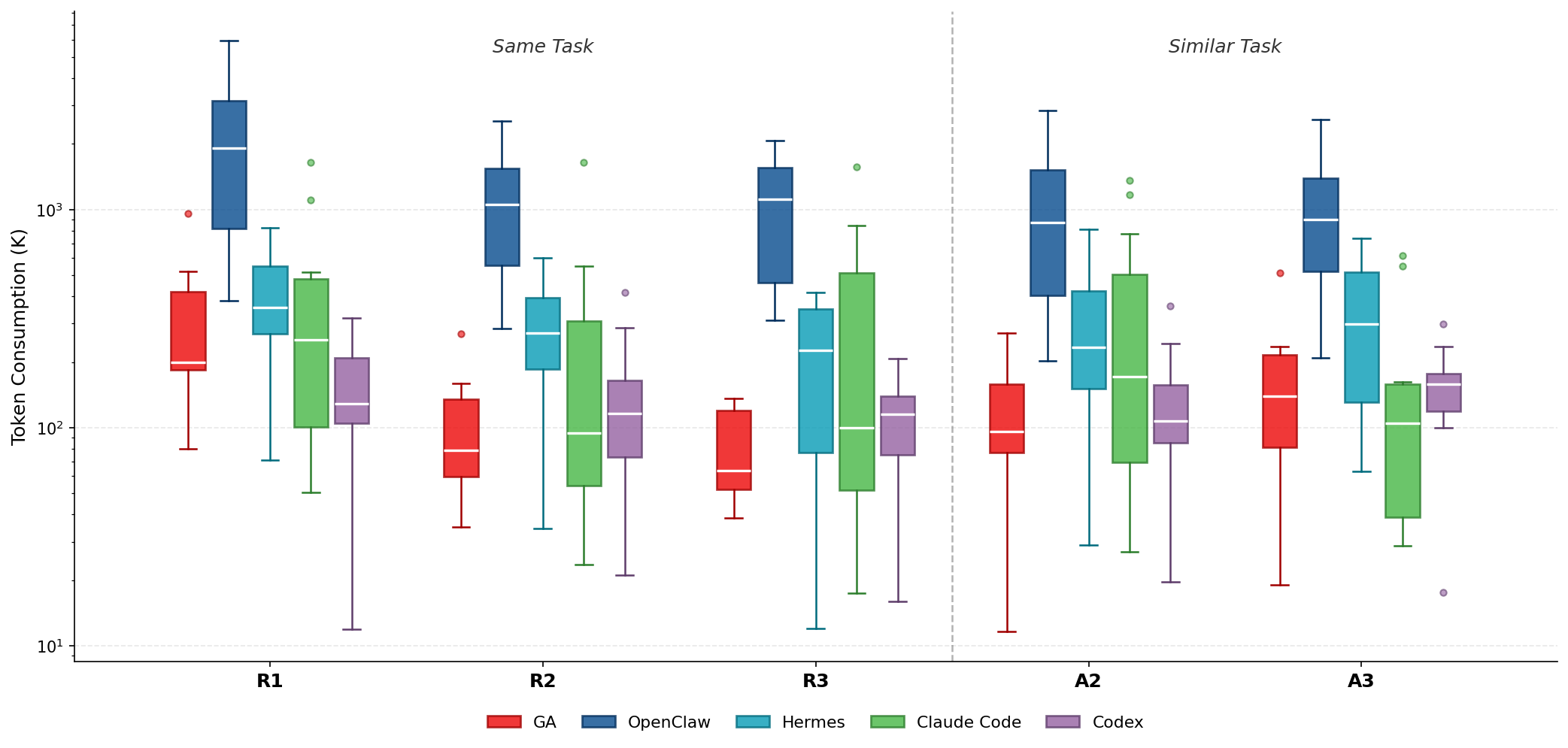}
\caption{Token consumption distribution across rounds on the Easy-Correlated sequence. The boxplot illustrates the spread and central tendency of each agent's token usage from R1 to R3.}
\label{fig:boxplot}
\end{figure*}

\section{Experiments}
 
\subsection{Experimental Setup}
 
To rigorously evaluate the capabilities of SEAs, we establish a controlled and reproducible experimental environment.
 
\paragraph{Evaluated Systems.} We select five representative SEAs for evaluation: GenericAgent (GA), OpenClaw\footnote{\url{https://github.com/openclaw/openclaw}}, Hermes\footnote{\url{https://github.com/NousResearch/hermes-agent}}, Claude Code, and Codex\footnote{\url{https://github.com/openai/codex}}. These five agents constitute the mainstream representative sample of the current SEAs.
 
\paragraph{Foundation Models.} The foundation model for each agent is configured according to task characteristics and system compatibility. Specific assignments are determined by each agent's native architecture and task requirements.
 
\paragraph{Execution Environment.} Given that SEAs possess system-level operational capabilities, all evaluations are strictly conducted within isolated containerized sandbox environments. This infrastructure ensures that agents can safely execute high-complexity, multi-step tasks without cross-contamination or host system damage. Memory and usage records are retained or cleared according to task requirements.
 
\subsection{Overall Performance Overview}
\label{sec:overall}

We first evaluate the five agents on the Easy-Correlated sequence. The foundation model for this sequence is uniformly set to Claude Opus 4.6; Codex adopts GPT-5.5 as an alternative due to architectural constraints. Table~\ref{tab:overall} presents the mean token consumption of each agent across sequence stages. Figure~\ref{fig:boxplot} displays the complete distributional characteristics\footnote{Fine-grained per-task token consumption data (with fail/timeout annotations) are available at \url{https://leaperovo.github.io/SEA-Eval/}. Due to time and token cost constraints, a 15-minute timeout mechanism was applied when testing Hermes, Claude Code, and Codex.}. Three preliminary observations warrant attention:

\begin{table}[t]
\centering
\caption{Mean token consumption (K tokens) and success rate of each agent on the Easy-Correlated sequence.}
\label{tab:overall}
\scriptsize
\begin{tabular}{lcccccc}
\toprule
\textbf{Agent} & \textbf{R1} & \textbf{R2} & \textbf{R3} & \textbf{A2} & \textbf{A3} & \textbf{SR} \\
\midrule
GA         & 198.7 & 78.8  & 63.6  & 96.0  & 139.8 & 95.0\% \\
OC   & 1905.1 & 1058.5 & 1118.7 & 868.7 & 897.5 & 96.7\% \\
Hermes     & 355.9 & 271.4 & 226.7 & 234.1 & 298.6 & 65.0\% \\
CC & 252.0 & 94.2  & 99.7  & 171.5 & 104.8 & 91.7\% \\
Codex      & 128.7 & 116.2 & 115.3 & 107.2 & 157.8 & 85.0\% \\
\bottomrule
\end{tabular}
\end{table}

\paragraph{(1) Token consumption varies widely across agents.} Although GA's R1-stage consumption is not the lowest overall, its aggregate token consumption is significantly lower than most agents. Notably, OpenClaw's consumption is extraordinarily high: OpenClaw's R1 mean (1905K) is 14.8$\times$ that of Codex (129K). Even after three learning rounds at the R3 stage, OpenClaw (1119K) remains 17.6$\times$ that of GA (64K). This disparity primarily stems from OpenClaw's lack of prompt compression, resulting in excessive tokens wasted on redundant context transmission.
 
\paragraph{(2) Evolutionary efficiency differs markedly across models.} GA exhibits a steep exponential decline (R1$\to$R2 accounts for 88\% of total optimization), Claude Code shows a fast-then-stable pattern (R1$\to$R2 decreases 63\%, R2$\to$R3 only 6\%), while Codex remains nearly flat (R1$\to$R3 decreases only 10\%). These differences reflect fundamentally distinct knowledge accumulation mechanisms.
 
\paragraph{(3) Success rate and token consumption exhibit a synergistic relationship.} Although the timeout mechanism affects SR, the token consumption data reveal that low-consumption agents (GA, Claude Code, Codex) maintain high success rates (85\%--95\%). This suggests that efficiency and reliability are not contradictory---streamlined execution strategies both conserve tokens and facilitate task completion. Hermes represents a notable counterexample: even on tasks not terminated by timeout, its token consumption is already significantly higher than GA, Claude Code, and Codex, indicating that its low SR is not solely attributable to timeout but reflects systemic inefficiency in execution.

\subsection{Evolutionary Capability Analysis}
\label{sec:evolution}
 
\begin{figure*}[t]
\centering
\includegraphics[width=0.95\textwidth]{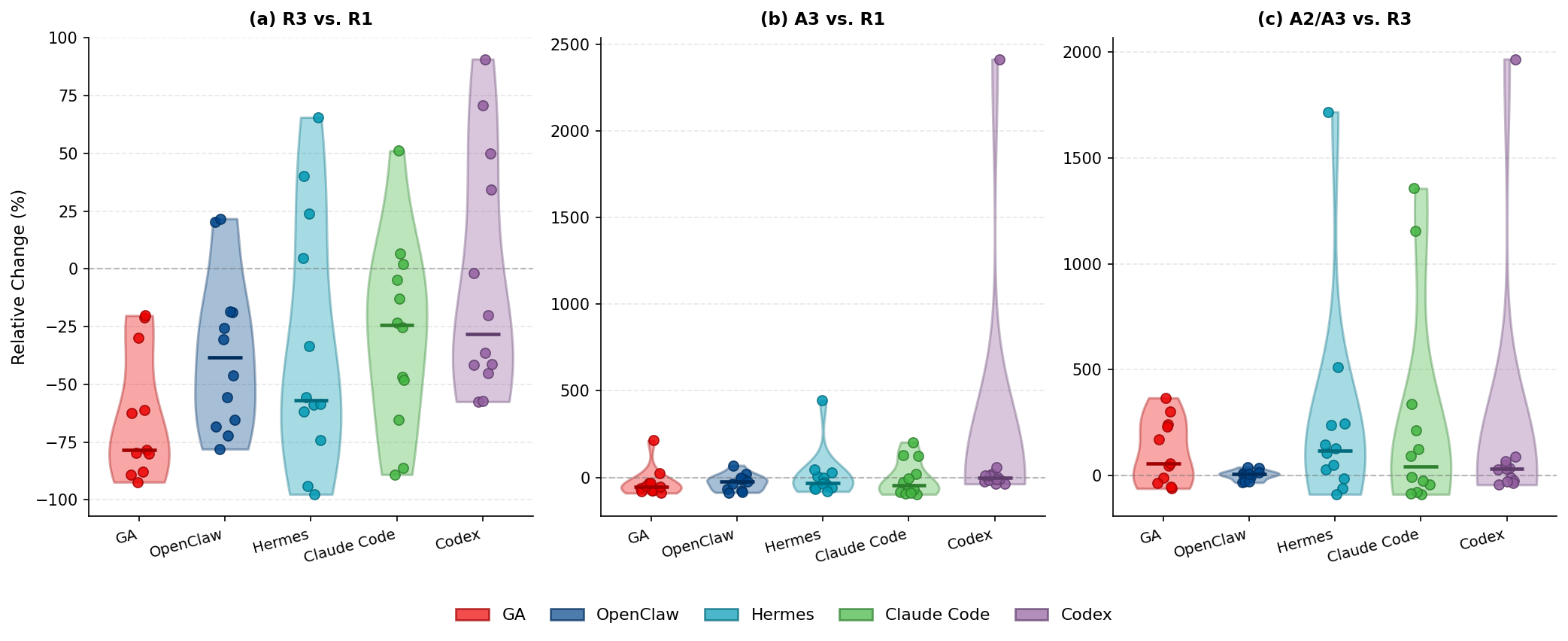}
\caption{Violin plots of token consumption changes across three conditions: (a) identical tasks (R3 vs.\ R1), (b) similar tasks (A2 vs.\ R1), and (c) similar tasks (A3 vs.\ R1). Negative values indicate efficiency improvement (token reduction).}
\label{fig:violin}
\end{figure*}

To analyze evolutionary efficiency under different conditions, we plot violin charts of token consumption changes on identical tasks (Figure~\ref{fig:violin}-a), as well as violin charts of consumption changes on similar tasks (Figure~\ref{fig:violin}-b and Figure~\ref{fig:violin}-c).
 
\paragraph{Finding 1: GA exhibits the most pronounced evolutionary efficiency on identical tasks.} Figure~\ref{fig:violin}-a shows that GA's R3 vs.\ R1 change rate median reaches $-78\%$, far exceeding other agents. Codex's learning effect is modest ($-28\%$), but this does not indicate weak learning capability---rather, Codex's R1-stage token consumption is already very low (129K), leaving limited optimization headroom. In other words, Codex's ``low learning rate'' is a natural consequence of its high initial efficiency.
 
\paragraph{Finding 2: Agents exhibit two fundamentally distinct evolutionary paradigms.} The first is the ``iterative optimization'' paradigm exemplified by GA---progressively compressing token consumption through multiple execution rounds, with learning curves showing clear declining trends. The second is the ``first-shot convergence'' paradigm exemplified by Claude Code and Codex---tending to reach near-optimal states on the first execution, with limited variation in subsequent rounds. Notably, both Claude Code and Codex benefit from prompt caching mechanisms~\cite{prompt_caching_2025,genericagent2025}, which substantially reduce token billing for repeated content, lending their consumption data a degree of opacity---ostensibly ``low consumption'' partially derives from cache hits rather than genuine strategy optimization. In contrast, OpenClaw exhibits negligible evolutionary amplitude, with its R3 vs.\ R1 change rate approaching zero, manifesting a ``pseudo-evolution'' characteristic: token consumption remains virtually unchanged across rounds, lacking genuine learning accumulation.
 
\paragraph{Finding 3: On transfer to similar tasks, agents differ little in transfer magnitude but substantially in stability.} Figure~\ref{fig:violin}-b and Figure~\ref{fig:violin}-c show that GA's A3 vs.\ R1 change rate median is $-55\%$, indicating that learning accumulated on identical tasks effectively transfers to similar tasks. Across agents, the transfer magnitude (median change rate) does not differ dramatically, yet the stability differs significantly: Hermes, Claude Code, and Codex each exhibit several abnormal task cases with sharply increased token consumption, resulting in highly volatile distributions on similar tasks. OpenClaw's apparent stability is highest, but this is not genuine transfer stability---rather, it stems from its ``pseudo-evolution'' characteristic: since token consumption consistently maintains high levels across all rounds, the change amplitude on similar tasks naturally remains flat, representing stagnation rather than robust transfer.
 
\begin{figure*}[t]
\centering
\includegraphics[width=0.95\linewidth]{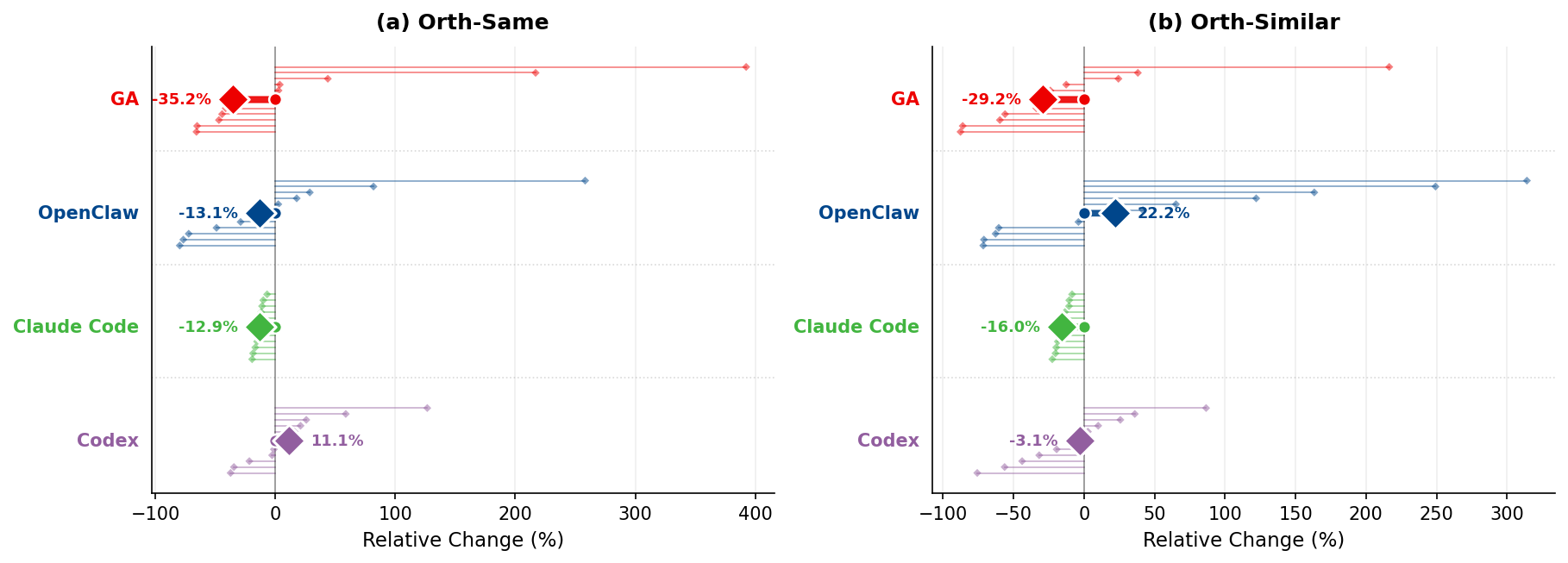}
\caption{Per-task token consumption trajectories under orthogonal interference. Diamonds indicate post-evolution values; circles indicate pre-evolution baselines. Median trajectories are highlighted with bold lines.}
\label{fig:orthogonal}
\end{figure*}

To further analyze the impact of unrelated tasks on agent performance, we conduct experiments on the Easy-Orth-Same and Easy-Orth-Similar datasets, with results presented in Figure~\ref{fig:orthogonal}.
 
\paragraph{Finding 4: GA retains its learning gains under cross-domain interference.} In orthogonal experiments, even when unrelated distractor tasks are interleaved between target task executions, GA's efficiency improvements on both identical and similar tasks are preserved (median change rates of $-35.2\%$ and $-29.2\%$, respectively, remaining in the negative range with considerable magnitude), indicating that its previously accumulated execution strategies are not disrupted by cross-domain knowledge. This suggests that GA's learning mechanism can effectively consolidate experience and reuse it after interference. The figure nonetheless reveals that GA's learning performance still exhibits some instability---Figure~\ref{fig:orthogonal} shows a small number of cases with increased token consumption, indicating that GA remains susceptible to interference from unrelated tasks, which we attribute to the insufficient stability of its memory retrieval mechanism.
 
\paragraph{Finding 5: After interference, OpenClaw's token consumption on similar tasks increases markedly.} Under the Orth-Similar condition, OpenClaw's median change rate is $+22.2\%$, in sharp contrast to the slight improvement ($-13.1\%$) under the Orth-Same condition. This again reflects OpenClaw's pseudo-evolution characteristic. Case analysis reveals that OpenClaw\footnote{This phenomenon is also present in Hermes. Note that Hermes's orthogonal experiments have not been fully completed; related conclusions are preliminary observations to be supplemented with complete data.} exhibits ``shortcut memorization'' behavior: during repeated execution of identical tasks, the agent may directly memorize operational answers, achieving extremely low token consumption. However, the cost of such mechanical memorization is that once task variants appear or context is disrupted, evolutionary performance deteriorates dramatically. Figure~\ref{fig:orthogonal} directly corroborates this fragility: OpenClaw is the only agent whose median change rate turns from negative to positive under the Orth-Similar condition.
 
\begin{table*}[t]
\centering
\caption{Stability metrics summary.}
\label{tab:stability}
\small
\begin{tabular}{lcccccc}
\toprule
\textbf{Agent} & \textbf{R2} & \textbf{R3} & \textbf{Stab\textsubscript{Id}} & \textbf{A2} & \textbf{A3} & \textbf{Stab\textsubscript{Sim}} \\
\midrule
GA         & 78.8  & 63.6   & $\mathbf{-19.3\%}$ & 96.0  & 139.8 & $+45.6\%$ \\
OpenClaw   & 1058.5 & 1118.7 & $+5.7\%$  & 868.7 & 897.5 & $\mathbf{+3.3\%}$ \\
Hermes     & 271.4 & 226.7  & $-16.5\%$ & 234.1 & 298.6 & $+27.6\%$ \\
Claude Code & 94.2  & 99.7   & $+5.8\%$  & 171.5 & 104.8 & $-38.9\%$ \\
Codex      & 116.2 & 115.3  & $-0.8\%$  & 107.2 & 157.8 & $+47.2\%$ \\
\bottomrule
\end{tabular}
\end{table*}
 
\subsection{Comprehensive Capability Profile}
 
To comprehensively evaluate the multi-dimensional capabilities of agents, we introduce task success rate (SR), token consumption (TC), multiple variants of $K_{\text{eff}}$, transfer performance ($K_{\text{trans}}$), and robustness ($K_{\text{eff}}(\text{Ret})$) as evaluation dimensions.\footnote{Detailed per-task token consumption data (with fail/timeout annotations) are available at \url{https://leaperovo.github.io/SEA-Eval/}.} Specifically, to assess learning effects when repeatedly executing the same task, we introduce $K_{\text{eff}}(\text{Evo})$ to measure overall improvement after multiple repetitions, and $K_{\text{eff}}(\text{Conv})$ to measure the speed at which learning occurs; to evaluate an agent's ability to retain acquired knowledge under orthogonal interference, we introduce $K_{\text{eff}}(\text{Stab-Id})$ for performance stability after same-task interference, and $K_{\text{eff}}(\text{Stab-Sim})$ for stability after similar-task interference; finally, to assess the balance between learning and forgetting, we compute $K_{\text{eff}}(\text{Ret})$, which comprehensively reflects the degree to which an agent retains old knowledge after acquiring new knowledge. Specifically, $K_{\text{eff}}(\text{Evo}) = (T_{R_1} - T_{R_3})/T_{R_1}$, $K_{\text{eff}}(\text{Conv}) = (T_{R_1} - T_{R_2})/T_{R_1}$, $K_{\text{trans}} = \frac{1}{2}[(T_{A_2} - T_{R_1})/T_{R_1} + (T_{A_3} - T_{R_1})/T_{R_1}]$, $K_{\text{eff}}(\text{Stab-Id}) = (T_{\text{post\_same}} - T_{\text{pre\_same}})/T_{\text{pre\_same}}$, $K_{\text{eff}}(\text{Stab-Sim}) = (T_{\text{post\_sim}} - T_{\text{pre\_sim}})/T_{\text{pre\_sim}}$, and $K_{\text{eff}}(\text{Ret}) = K_{\text{eff}}(\text{Conv})/K_{\text{eff}}(\text{Stab-Id})$.
 
 
Here, $K_{\text{eff}}(\text{Evo})$ and $K_{\text{eff}}(\text{Conv})$ measure learning efficiency (higher values indicate greater token consumption reduction), $K_{\text{trans}}$ measures cross-task transfer capability (negative values indicate effective transfer), $K_{\text{eff}}(\text{Stab-Id})$ and $K_{\text{eff}}(\text{Stab-Sim})$ measure post-interference stability (values near zero indicate stability), and $K_{\text{eff}}(\text{Ret})$ comprehensively reflects learning retention rate (larger values indicate faster learning relative to forgetting).
 
\begin{figure*}[t]
\centering
\includegraphics[width=0.45\textwidth]{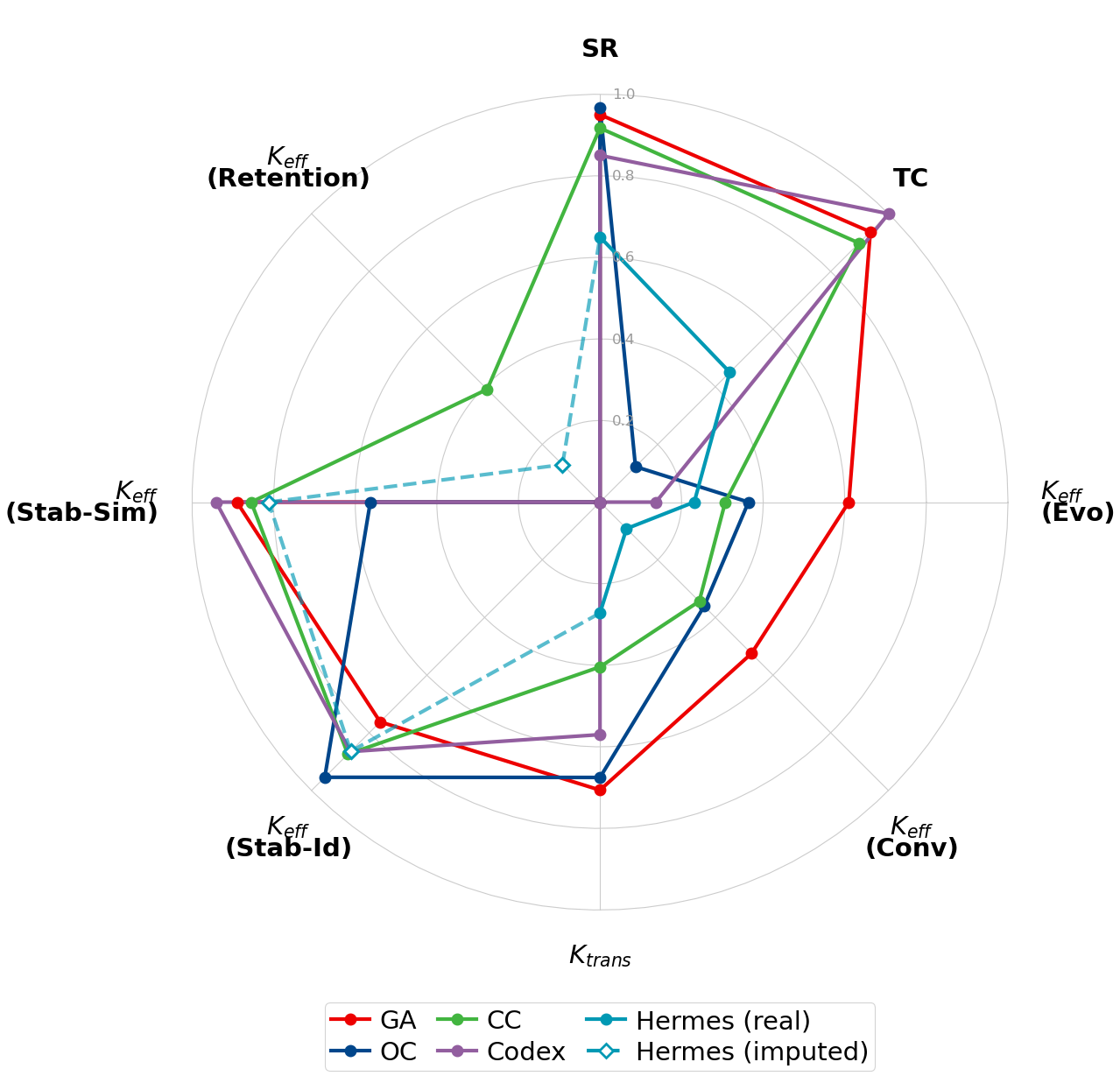}
\hfill
\includegraphics[width=0.45\textwidth]{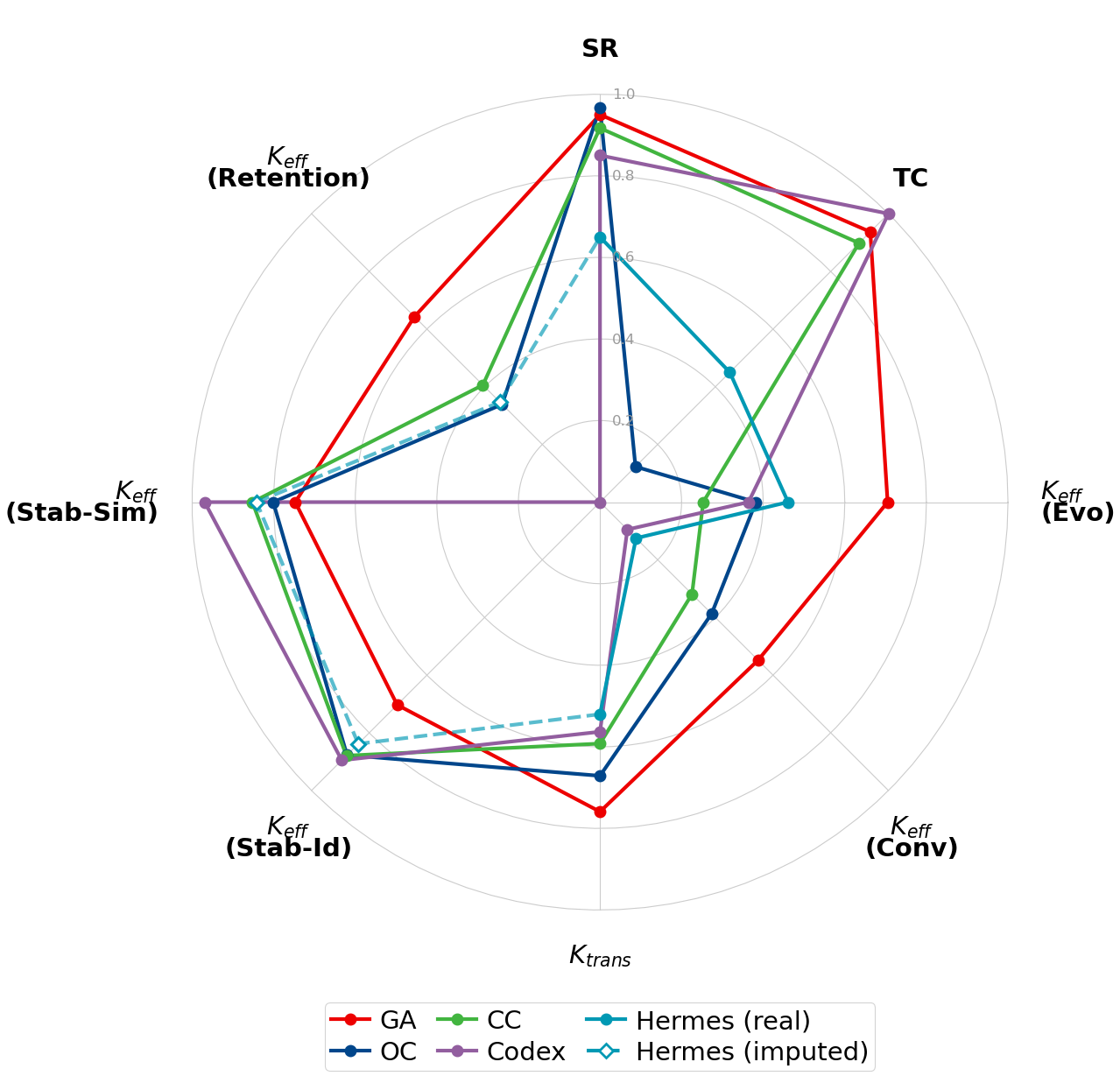}
\caption{Radar charts of comprehensive capability profiles using mean (left) and median (right) aggregation across 8 evaluation dimensions.}
\label{fig:radar}
\end{figure*}

We normalize all metrics to the $[0, 1]$ interval\footnote{Detailed normalization methodology is provided in Appendix~\ref{app:normalization}.} and generate radar charts (Figure~\ref{fig:radar}) using both median and mean as per-task aggregation functions. Table~\ref{tab:comprehensive} presents the 8-dimensional composite scores (AVG) under both aggregation methods:
 
\begin{table}[t]
\centering
\caption{Comprehensive capability scores.}
\label{tab:comprehensive}
\small
\begin{tabular}{lcc}
\toprule
\textbf{Agent} & \textbf{Median-AVG} & \textbf{Mean-AVG} \\
\midrule
GA         & \textbf{0.737} & \textbf{0.668} \\
Claude Code & 0.637 & 0.621 \\
Codex      & 0.591 & 0.560 \\
OpenClaw   & 0.565 & 0.483 \\
Hermes     & 0.524 & 0.431 \\
\bottomrule
\end{tabular}
\end{table}
 
The rankings are highly consistent across both aggregation methods: GA $>$ Claude Code $>$ Codex $>$ OpenClaw $>$ Hermes. GA ranks first with significant margins under both perspectives, with its core advantage deriving from comprehensive leadership across the learning dimensions ($K_{\text{eff}}(\text{Evo})$/$K_{\text{eff}}(\text{Conv})$/$K_{\text{trans}}$/$K_{\text{eff}}(\text{Ret})$). Claude Code consistently occupies second place, approaching Codex in efficiency and stability dimensions while retaining moderate learning capability.
 
The primary divergence between Median and Mean appears in GA's $K_{\text{eff}}(\text{Ret})$ dimension: Median yields 0.643 (knowledge retention is good for most tasks), while Mean drops to 0 (a few extreme degradation tasks pull the average into negative territory, clipped to 0), corroborating the high-variance characteristic of GA identified in Section~\ref{sec:evolution}.
 
Based on radar chart morphology and numerical distributions, we categorize the five agents into two archetypal patterns:
 
\paragraph{(1) Efficient Learning Type (GA).} GA leads comprehensively across the four learning dimensions ($K_{\text{eff}}(\text{Evo})$, $K_{\text{eff}}(\text{Conv})$, $K_{\text{trans}}$, $K_{\text{eff}}(\text{Ret})$), demonstrating the strongest experience accumulation and strategy optimization capabilities. In terms of raw measurements, GA achieves a 68.0\% overall efficiency gain on identical tasks (R1$\to$R3) and a 60.3\% convergence rate (R1$\to$R2), both ranking first among all agents. However, GA's memory retrieval mechanism still has room for improvement: under orthogonal interference, its normalized $K_{\text{eff}}(\text{Stab-Id})$ score is 0.648---the lowest among all agents---corresponding to a raw change rate of $-19.3\%$, a fluctuation markedly larger than that of OpenClaw ($+5.7\%$), Claude Code ($+5.8\%$), and Codex ($-0.8\%$). This indicates that while GA's learning depth is strong, the instability of its memory management constrains its cross-task generalization and interference resistance.
 
\paragraph{(2) Stable Enhancement Type (Claude Code, Codex).} Claude Code and Codex benefit from caching advantages in their self-deployed architectures, achieving the highest normalized TC scores among all agents (0.898 and 1.000, respectively), which makes their memory management inherently more efficient---reduced context burden translates to clearer state maintenance and lower interference risk. Claude Code maintains high stability (normalized $K_{\text{eff}}(\text{Stab-Id})=0.871$, $K_{\text{eff}}(\text{Stab-Sim})=0.840$) while preserving appreciable learning capability ($K_{\text{eff}}(\text{Evo})=0.253$, $K_{\text{eff}}(\text{Ret})=0.406$); in terms of raw measurements, it is the only agent that continues optimization on similar tasks after orthogonal interference (Stab-Sim raw change rate of $-38.9\%$), making it currently the only agent that achieves a favorable balance in the ``learning--stability'' tradeoff. Codex pursues an extreme efficiency route, trading minimal cost (TC$=1.000$) and maximum consistency ($K_{\text{eff}}(\text{Stab-Sim})=0.969$) for near-complete abandonment of learning capability ($K_{\text{eff}}(\text{Conv})=0.095$, $K_{\text{eff}}(\text{Ret})=0.000$).
 
These two patterns reveal the core tradeoff in current agent design: the tension between learning depth and execution stability. GA's aggressive learning strategy yields the highest evolutionary returns, but its dependence on memory retrieval quality leads to instability in transfer and interference-resistance scenarios; Claude Code's conservative strategy reduces memory management complexity through structured caching, achieving more reliable overall performance at the cost of partial learning depth.
\section{Conclusion}

This paper presents SEA-Eval, a systematic evaluation framework for Self-Evolving Agents that integrates intra-task execution reliability and long-term evolutionary performance. We contribute a formal definition of the SEA paradigm grounded in digital embodiment and continuous cross-task evolution, and establish the Evolutionary Flywheel as its structural necessity. Building on this theoretical foundation, SEA-Eval moves beyond performance snapshots to measure evolutionary trajectories directly, supported by a multidimensional metric system centered on Success Rate and Token Consumption, and a standardized dataset organized into sequential task streams for observing agent behavior in complex, non-stationary environments.
Empirical results reveal a fundamental gap between apparent capability and genuine evolution: frameworks achieving identical success rates exhibit up to approximately 30× divergence in token consumption, and only architectures equipped with effective experience distillation demonstrate the monotonic convergence that distinguishes true evolution from pseudo-evolution. These findings suggest that self-evolving agents, as a form of continuously adaptive digital entities, represent a meaningful step toward more capable and efficient autonomous systems. Challenges remain in areas such as fully autonomous bootstrapping, subjective preference alignment, and systemic governance, and the evaluation principles established in this work provide a concrete foundation for addressing them in future research.

\newpage

\section{Discussion and Future Work}

SEA-Eval establishes a rigorous foundation for evaluating the evolutionary 
dynamics of single-agent systems across intra-task reliability and long-term 
performance dimensions. However, the SEA paradigm as formalized in 
Section~\ref{sec:sea} opens several directions that extend beyond the current 
benchmark's scope. We organize these directions across three frontiers: the 
capability frontier, concerning what SEAs can do and how efficiently they do 
it; the evaluation frontier, concerning what aspects of SEA behavior remain 
difficult to measure; and the deployment frontier, concerning how SEAs can be 
safely and broadly instantiated in the real world.

\subsection{Expanding the Capability Frontier}

\subsubsection{Toward Fully Autonomous Bootstrapping}

Current SEA deployments rely on human-provided task specifications to 
initialize the first execution cycle, from which the Evolutionary Flywheel 
begins distillation. A fully autonomous SEA must eventually initiate its own 
evolutionary cycle without human scaffolding---identifying performance 
bottlenecks, generating synthetic experiences, and triggering self-iteration 
in the absence of explicit external prompts. Solving this cold-start problem 
is a prerequisite for SEAs to achieve genuine generalization in unknown 
domains, and represents a capability dimension not currently assessed by 
SEA-Eval.

\subsubsection{Balancing Reasoning Depth and Execution Efficiency}

The current SEA architecture relies on large language models as the cognitive 
core of the Cognition Hub, whose inference process introduces non-negligible 
latency overhead. For tasks requiring immediate feedback---such as real-time 
interaction, high-frequency operations, or latency-sensitive workflows---the 
output efficiency of current models falls short of practical deployment 
demands. This gap is particularly pronounced in industrial-grade scenarios, 
where model inference latency and task execution requirements diverge 
significantly, limiting the applicability of SEAs in such contexts. One 
promising direction is heterogeneous multi-model orchestration: delegating 
latency-sensitive, high-frequency execution tasks to lightweight models while 
reserving large models for offline experience distillation and strategic 
planning. How to appropriately delineate the responsibility boundary between 
these two classes of models, and how to introduce latency-aware evaluation 
dimensions into the SEA-Eval framework, remain open problems for future work.

\subsubsection{Toward Collective Evolution: Multi-agent Co-evolution}

The $\mathcal{SEA}$ formalization characterizes the evolutionary dynamics of a 
single agent operating within a private memory state $\mathcal{M}$. However, 
the bounded capacity of a single agent's $\mathcal{M}$ imposes an upper limit 
on achievable evolutionary gains, and expanding productivity beyond this 
ceiling naturally motivates multi-agent coordination. When multiple SEAs 
operate within shared or overlapping environments, new questions emerge: how 
distilled knowledge in $\mathcal{M}_p$ should be selectively shared across 
agents without introducing noise from agent-specific execution contexts; how 
coupled Evolutionary Flywheels interact to produce system-level evolutionary 
gains that may exceed the sum of individual contributions; and how functional 
specialization emerges across agents over extended co-deployment. These 
questions require both theoretical extensions to the $\mathcal{SEA}$ 
formalization and new system-level evaluation primitives beyond the individual 
$K_{\text{eff}}$ and $K_{SR}$ metrics defined in SEA-Eval.

\subsection{Expanding the Evaluation Frontier}

\subsubsection{Subjective Preference and Deep Alignment}

SEA-Eval operationalizes user alignment through $K_{\text{align}}$, which 
tracks the longitudinal reduction in user intervention rate as a behavioral 
proxy for preference internalization. However, this metric captures only the 
observable behavioral layer of alignment. A large class of alignment 
phenomena---such as the aesthetic quality of a generated presentation or the 
stylistic appropriateness of produced content---involve subjective, 
individualized judgments whose ground truth cannot be formalized as 
intervention frequencies or task success labels. Future research should 
explore evaluation paradigms that move beyond behavioral proxies toward richer 
preference elicitation methods, capturing the convergence between an agent's 
evolved strategies and a user's latent aesthetic and functional intent.

\subsection{Expanding the Deployment Frontier}

\subsubsection{Safety and Systemic Governance}

As SEAs autonomously expand $\mathcal{A}$ and update $\mathcal{M}$, safety 
and governance become non-negotiable deployment requirements. Two structural 
insights frame this challenge. First, capability-permission parity: broader 
system permissions enhance evolutionary potential but proportionally scale the 
risk profile of autonomous self-modification, and no principled framework 
currently exists for calibrating permission scope to evolutionary stage. 
Second, safety as system engineering: agent-level safeguards are inherently 
vulnerable to the same distillation failures that produce pseudo-evolution, 
making it necessary to enforce evolutionary boundaries through underlying 
permission management frameworks and sandboxed execution environments rather 
than relying on agent-level constraints alone. Future research should 
investigate how the Evolutionary Flywheel can be integrated into governed 
scaffolding architectures that keep evolved assets in $\mathcal{M}$ auditable 
and controllable within mission-critical workflows.

\subsubsection{From Digital Embodiment to Embodied AI}

The SEA architecture is built around a modality-agnostic closed loop: the 
Execution Hub perceives environmental states, acts upon them, and returns 
feedback; the Cognition Hub distills the resulting trajectory into persistent 
memory updates via $\Phi$ and $\Psi$. This loop is instantiated in digital 
environments through screen pixels and system logs feeding into keyboard and 
mouse actions, but its structural logic is isomorphic to the 
perception-action-feedback cycles of physical robotic systems. Since 
$\Phi: \tau \rightarrow \mathcal{M}$ and 
$\Psi: \mathcal{M}_t \rightarrow \mathcal{M}_{t+1}$ are modality-independent 
by definition, the transition from digital to physical embodiment is primarily 
a challenge of modality adaptation rather than architectural reinvention. 
Digital environments therefore offer a high-fidelity, low-cost testbed for 
developing the evolutionary mechanisms that physical robotic systems will 
ultimately require.

\section{Ethics Statement}
\label{sec:ethics}

\paragraph{Data Source and Privacy}
All tasks in SEA-Eval are constructed by human experts based on publicly available web resources and local system operation scenarios. No personal privacy data or user-sensitive information is involved. All web automation tasks target publicly accessible websites, and all local system operation tasks are executed within isolated sandbox environments, ensuring that no real user data is collected or stored during evaluation. The benchmark dataset and code will be publicly released upon publication to support reproducibility.

\paragraph{Experimental Safety}
Since SEA operates with autonomous system-level capabilities including code execution, file manipulation, and environment configuration, all experiments in this paper are conducted strictly within containerized sandbox environments. This ensures that agent operations produce no unintended effects on host systems or external environments.

\paragraph{AI-Assisted Writing and Figure Generation}
This paper used Claude Sonnet 4.6 for language polishing and expression refinement during the writing process. The framework diagrams are generated with the assistance of Gemini 3. All research design, experimental implementation, data analysis, and core arguments were completed independently by the authors. The use of these AI tools was limited to improving the fluency and academic conventions of English expression and the visual presentation of diagrams, and did not involve substantive contributions to the research content.

\bibliography{custom}

\appendix

\section{Normalization Methodology}
\label{app:normalization}

To enable cross-metric comparison on radar charts, we normalize all raw metrics to the $[0, 1]$ interval. Since different metrics have distinct physical semantics and value ranges, we adopt metric-specific normalization strategies (shown in Table~\ref{tab:normalization}):

\begin{table*}[h]
\centering
\caption{Normalization strategies for each metric dimension.}
\label{tab:normalization}
\small
\begin{tabular}{lll}
\toprule
\textbf{Metric} & \textbf{Formula} & \textbf{Rationale} \\
\midrule
SR & $\text{raw}$ (already in $[0,1]$) & Direct use \\
TC & $\text{clip}(1 - \text{raw}/\text{max}, 0, 1)$ & Lower is better; invert \\
$K_{\text{eff}}$(Evo) & $\text{clip}(\text{raw}, 0, 1)$ & Positive = improvement \\
$K_{\text{eff}}$(Conv) & $\text{clip}(\text{raw}, 0, 1)$ & Positive = fast learning \\
$K_{\text{trans}}$ & $\text{clip}(-\text{raw}/2, 0, 1)$ & Negative raw = effective transfer \\
$K_{\text{eff}}$(Stab-Id) & $\exp(-|\text{raw}|)$ & Near-zero = stable \\
$K_{\text{eff}}$(Stab-Sim) & $\exp(-|\text{raw}|)$ & Near-zero = stable \\
$K_{\text{eff}}$(Ret) & $\text{clip}(\text{raw}, 0, 1)$ & Positive = good retention \\
\bottomrule
\end{tabular}
\end{table*}

\paragraph{Design Rationale.} For stability metrics (Stab-Id/Stab-Sim), we adopt exponential decay rather than linear mapping. The motivation is that small fluctuations in token consumption (e.g., $<20\%$) are typically acceptable in practice and should not incur linear penalties, whereas large fluctuations (e.g., $>50\%$) should be substantially penalized. The exponential function $\exp(-|x|)$ exhibits a gentle gradient near the origin and rapid decay away from it, precisely matching this physical intuition.

\section{Functional Boundary Between $\Phi$ and $\Psi$}
\label{appendix:phi_psi}

The formal definition of $\mathcal{SEA}$ in Section~\ref{sec:characteristics} introduces two functions, $\Phi: \tau \rightarrow \Delta\mathcal{M}$ and $\Psi: \mathcal{M}_t \times \Delta\mathcal{M} \rightarrow \mathcal{M}_{t+1}$, whose functional boundary warrants explicit clarification, as conflating their roles leads to an inconsistent interpretation of the SEA's evolutionary mechanism.

\paragraph{The Role of $\Phi$: Experience Distillation.}
$\Phi$ operates on the raw interaction trajectory $\tau$ produced by a single task execution and is responsible for extracting structured knowledge from it. Specifically, $\Phi$ performs causal attribution to identify the critical decision subsequence $\tau^* \subset \tau$ that produced the observed outcome, and abstracts $\tau^*$ into a structured knowledge update $\Delta\mathcal{M}$. This update is decomposed across the three memory layers: successful execution paths are abstracted into parameterized strategies destined for $\mathcal{M}_p$; failure attributions and diagnostic records are destined for $\mathcal{M}_e$; and inferred environmental constraints and user preference signals are destined for $\mathcal{M}_d$. Critically, $\Phi$ operates at the level of a single trajectory and produces a candidate update---it does not directly modify $\mathcal{M}$. The quality of $\Phi$'s output is determined by its abstraction capacity: a high-quality $\Phi$ lifts instance-specific execution details to the level of generalizable strategies, while a low-quality $\Phi$ produces updates that are overfitted to the specific trajectory and fail to transfer to structurally similar future tasks.

\paragraph{The Role of $\Psi$: Selective Memory Integration.}
$\Psi$ operates on the current memory state $\mathcal{M}_t$ and the candidate update $\Delta\mathcal{M}$ produced by $\Phi$, and is responsible for determining which components of $\Delta\mathcal{M}$ are committed to the persistent memory state $\mathcal{M}_{t+1}$. This selectivity serves two purposes. First, it preserves evolutionary stability by suppressing updates that fail to meet generalizability criteria---preventing instance-specific knowledge from polluting $\mathcal{M}_p$ and degrading the quality of future retrievals. Second, it manages memory capacity by suppressing redundant updates that duplicate existing entries in $\mathcal{M}_t$, preventing unbounded growth of $\mathcal{M}$ that would increase retrieval overhead over time. $\Psi$ therefore acts as a gating mechanism between the distillation process and the persistent memory state, ensuring that $\mathcal{M}$ grows in quality rather than merely in cardinality.

\paragraph{The Interaction Between $\Phi$ and $\Psi$.}
The two functions are sequentially coupled: $\Phi$ is applied first to produce $\Delta\mathcal{M}$, which is then passed to $\Psi$ for selective integration. This sequential coupling has an important implication for failure mode analysis: the quality of $\mathcal{M}_{t+1}$ depends jointly on the abstraction quality of $\Phi$ and the selectivity precision of $\Psi$. A high-quality $\Phi$ paired with an overly conservative $\Psi$ will suppress genuinely generalizable updates, stunting evolutionary growth; a low-quality $\Phi$ paired with an overly permissive $\Psi$ will admit instance-specific noise into $\mathcal{M}_p$, degrading retrieval precision. The evolutionary flywheel operates optimally when both functions are jointly calibrated---a constraint that has direct implications for the design of SEA architectures and the diagnosis of pseudo-evolution, as discussed in Appendix~\ref{appendix:failure}.

\paragraph{Degenerate Case: Episodic Agents.}
In episodic agents, both $\Phi$ and $\Psi$ degenerate to null operations. $\Phi$ produces an empty update $\Delta\mathcal{M} = \emptyset$ at every task boundary, and $\Psi$ trivially maps $\mathcal{M}_t$ to a reset state $\mathcal{M}_{t+1} = \emptyset$ regardless of any input. The SEA definition formally demarcates this boundary by requiring both functions to be non-trivial: any system in which $\Phi$ consistently produces $\Delta\mathcal{M} = \emptyset$ or $\Psi$ consistently discards all updates is, by definition, an episodic agent, regardless of the sophistication of its within-task reasoning.

\section{Diagnostic Analysis of Evolutionary Flywheel Failure Modes}
\label{appendix:failure}

Section~\ref{sec:flywheel} identifies two theoretically distinct failure modes---distillation failure and retrieval failure---through which the Evolutionary Flywheel degrades into pseudo-evolution. As noted there, both failure modes produce an identical primary behavioral signature: execution cost fails to decrease across sequential task variants and scales linearly with task frequency rather than converging. This behavioral equivalence at the observation level raises a diagnostic question: given an agent exhibiting pseudo-evolution, how can one determine which failure mode is responsible? This appendix provides a formal characterization of each failure mode and discusses how they can be distinguished through auxiliary execution traces.

\subsection{Formal Characterization of Failure Modes}

\paragraph{Distillation Failure.}
Distillation failure occurs when $\Phi$ cannot perform effective causal attribution on the raw interaction trajectory $\tau$. Formally, let $\tau^*$ denote the critical decision subsequence causally responsible for the observed outcome. Distillation failure arises when $\Phi$ either (1) fails to identify $\tau^*$ within $\tau$---producing a memory update $\Delta\mathcal{M}$ that encodes instance-specific details rather than generalizable strategies---or (2) produces a $\Delta\mathcal{M}$ of insufficient abstraction quality such that $\Psi$ correctly commits it to $\mathcal{M}_{t+1}$, yet the committed knowledge fails to transfer to structurally similar future tasks. In both cases, $\mathcal{M}_p$ remains effectively inert: its cardinality may increase across task boundaries, but its entries do not reduce execution overhead in subsequent tasks. Distillation failure can arise from two sources: trajectory noise, in which the raw trajectory $\tau$ contains a high proportion of irrelevant or contradictory actions that obscure the causal structure; and insufficient abstraction capacity, in which the Cognition Hub lacks the reasoning depth to lift instance-specific execution details to the level of reusable parameterized strategies.

\paragraph{Retrieval Failure.}
Retrieval failure occurs when the Execution Hub cannot identify structural similarity between a new task and existing entries in $\mathcal{M}$, despite $\mathcal{M}_p$ containing relevant and generalizable strategies. Formally, let $m^* \in \mathcal{M}_p$ denote the memory entry most relevant to the current task. Retrieval failure arises when the similarity function used to query $\mathcal{M}$ assigns $m^*$ a rank below the retrieval threshold, causing the Execution Hub to default to zero-shot reasoning and bypass the flywheel entirely. Retrieval failure is particularly likely under environmental drift: when the surface-level features of a new task diverge from those of previously encountered tasks---due to interface changes, domain shifts, or linguistic reformulations---shallow similarity metrics fail to recognize the underlying structural correspondence. Unlike distillation failure, retrieval failure does not imply any deficiency in the quality of $\mathcal{M}_p$; the knowledge is present and generalizable, but inaccessible to the retrieval mechanism under the current task presentation.

\subsection{Diagnostic Distinction via Auxiliary Execution Traces}

Although both failure modes produce identical primary signatures in terms of token consumption trajectories, they leave distinguishable traces in auxiliary execution data. We describe two diagnostic approaches that can be applied when detailed execution logs are available.

\paragraph{Memory Update Monitoring for Distillation Failure.}
Distillation failure can be diagnosed by inspecting the content and growth trajectory of $\mathcal{M}_p$ across sequential task executions. Specifically, if $\mathcal{M}_p$ grows in cardinality across task boundaries yet execution overhead does not decrease, this is consistent with distillation failure: new entries are being committed to $\mathcal{M}_p$, but they lack sufficient generalizability to influence future execution paths. A more direct diagnostic signal is the \textit{strategy reuse rate}: the proportion of execution steps in task $t$ that invoke a strategy retrieved from $\mathcal{M}_p$ rather than generated through zero-shot reasoning. In distillation failure, the strategy reuse rate remains near zero even as $|\mathcal{M}_p|$ grows, indicating that committed entries are not being matched to---or are being matched but failing to reduce the reasoning burden of---subsequent tasks. This distinguishes distillation failure from retrieval failure, in which the strategy reuse rate is also near zero but for a different reason: relevant entries exist in $\mathcal{M}_p$ but are not retrieved.

\paragraph{Retrieval Trace Analysis for Retrieval Failure.}
Retrieval failure can be diagnosed by inspecting the retrieval traces produced during task execution---specifically, the ranked list of $\mathcal{M}_p$ entries returned by the similarity query for each task. If the retrieval trace shows that $\mathcal{M}_p$ contains entries that are structurally relevant to the current task (as assessed by post-hoc human or automated analysis) yet these entries receive low similarity scores and are not surfaced to the Execution Hub, this constitutes direct evidence of retrieval failure. A complementary diagnostic is the \textit{retrieval precision under perturbation}: given a task $t$ and a known-relevant entry $m^* \in \mathcal{M}_p$, retrieval failure is confirmed if small surface-level reformulations of $t$---that preserve its structural execution requirements---produce large changes in the rank of $m^*$ in the retrieval output. This sensitivity to surface-level variation is the defining characteristic of retrieval failure under environmental drift.

\subsection{Implications for SEA-Eval}

The diagnostic framework described above has direct implications for the interpretation of SEA-Eval results. The primary metric of token consumption trajectory, as defined in Section~\ref{sec:metrics}, serves as the first-order signal for detecting pseudo-evolution. When pseudo-evolution is detected, the auxiliary metrics defined in Section~\ref{sec:metrics}---in particular execution steps $S$ and self-correction frequency $N_{\text{fix}}$---provide partial diagnostic information: a high $N_{\text{fix}}$ with stable $S$ is more consistent with distillation failure (the agent is attempting execution but failing to leverage prior experience), while a low $N_{\text{fix}}$ with high $T$ is more consistent with retrieval failure (the agent is engaging in extensive zero-shot reasoning without attempting to retrieve from $\mathcal{M}$). Full diagnostic resolution, however, requires access to internal memory update logs and retrieval traces, which are available in controlled experimental settings but may not be accessible in black-box agent evaluations. SEA-Eval therefore treats failure mode diagnosis as a best-effort auxiliary analysis rather than a primary evaluation requirement, and encourages future benchmark extensions to incorporate standardized memory and retrieval logging interfaces to enable systematic diagnostic evaluation.

\end{document}